\documentclass{article}

\usepackage{acra}

\usepackage{graphicx}
\usepackage{amsmath}
\usepackage{mathtools}
\usepackage{pifont}
\usepackage{algorithm}
\usepackage{algpseudocode}
\usepackage{eqparbox}
\usepackage{amssymb}
\usepackage{multicol}
\usepackage{multirow}
\usepackage[hidelinks]{hyperref}
\usepackage{float} 
\usepackage[dvipsnames]{xcolor}
\usepackage{diagbox}

\newcommand{\langt}{\mathbf{l}_t}
\newcommand{\vist}{\mathbf{o}_t}

\newcommand{\lang}{\mathbf{l}_t}
\newcommand{\vis}{\mathbf{o}_t}
\newcommand{\citeay}[1]{\cite{#1}}

\newcommand{\policy}{\pi}

\usepackage{tabularx}
\usepackage{multirow}
\title{ROSO: Improving Robotic Policy Inference via Synthetic Observations}

\author{Yusuke Miyashita, Dimitris Gahtidis, Colin La, Jeremy Rabinowicz, Jürgen Leitner\\Monash University\\https://yusuke710.github.io/roso.github.io/}

\begin{document}

\maketitle

\begin{abstract}
In this paper, we propose the use of generative artificial intelligence (AI) to improve zero-shot performance of a pre-trained policy by altering observations during inference. 
Modern robotic systems, powered by advanced neural networks, have demonstrated remarkable capabilities on pre-trained tasks. However, generalising and adapting to new objects and environments is challenging, and fine-tuning visuomotor policies is time-consuming. 
To overcome these issues we propose \textbf{Ro}botic Policy Inference via \textbf{S}ynthetic \textbf{O}bservations (\textbf{ROSO}). 
 
ROSO uses Stable Diffusion to pre-process a robot's observation of novel objects during inference time to fit within its distribution of observations of the pre-trained policies.
This novel paradigm allows us to transfer learned knowledge from known tasks to previously unseen scenarios, enhancing the robot's adaptability without requiring lengthy fine-tuning. 
Our experiments show that incorporating generative AI into robotic inference significantly improves successful outcomes, finishing up to 57\% of tasks otherwise unsuccessful with the pre-trained policy.

\end{abstract}

\begin{figure}[htp]
    \centering
    \includegraphics[width=.97\linewidth]{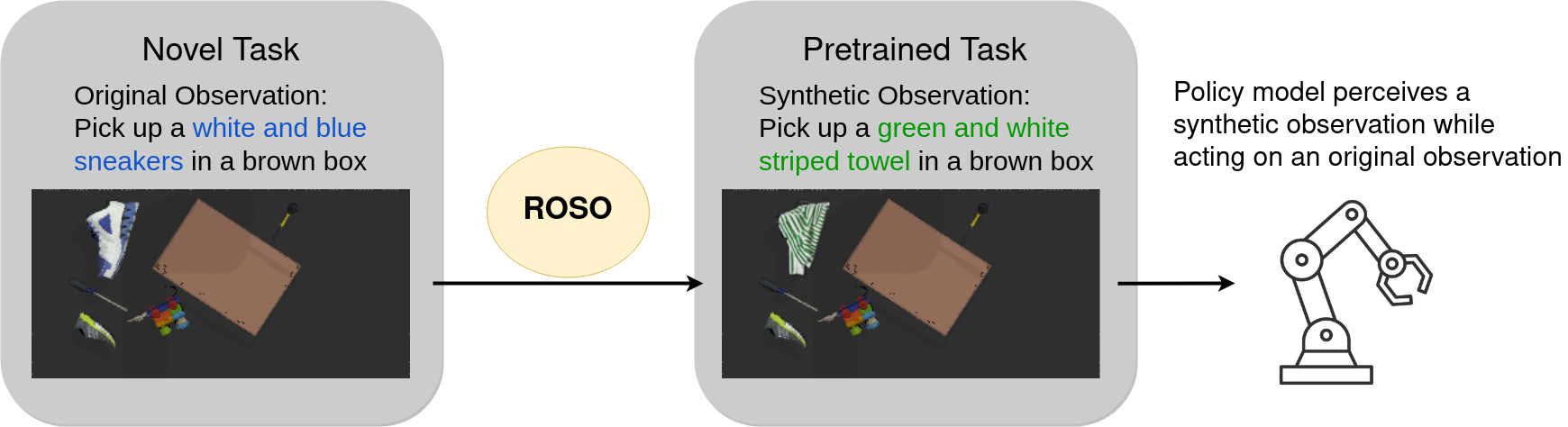}
    \vspace{-4mm}
    \caption{Given a pretrained robotic policy model, our synthetic observation pipeline (ROSO) transforms novel tasks into pretrained tasks. ROSO  
    yields successful policy execution in otherwise unsuccessful scenarios, without the need to retrain.} 
    \label{fig:vis-abstract}
\end{figure}

\section{Introduction}
Advances in neural networks
have enabled performance improvements of modern robotic systems on a set of pre-trained tasks. For example, recent publications employing text prompts have improved networks trained for pick-and-place tasks that are capable of picking, placing and
manipulating a range of objects in different environments \citeay{cliport}. Since training in multiple environments with
multiple objects is computationally expensive, these networks are only trained on specific objects and static environments. Similarly, fine-tuning a neural network is difficult, especially on models trained on internet-scale data such as CLIP
\cite{clip}. Therefore, it is desirable to train a robot in a specific
environment, have it master a task involving a small set of objects, and then find a way to transfer its learning to unseen environments and objects.

Recent work accentuates the potential to improve robotic policies through internet-scale pre-training~\citeay{rt2}. Leveraging visual-language models (VLM) yields remarkable enhancements in results and enables the learning of a single multi-task policy with the intent of increased performance in novel environments or
new objects (also referred to as ``zero-shot performance''). 
Inspired by recent work on augmenting the training set with generative methods~\citeay{rosie}, we propose to improve a policy's performance by manipulating a robot’s perception using generative AI in inference time rather than training (Figure~\ref{fig:vis-abstract}).

We show that shifting the out-of-distribution observation into the known (pre-trained) domain -- in our case, we replace a novel object and prompt with a known object -- enables the use of the pre-trained policy in those unseen scenarios without the need for re-training or fine-tuning. 
Our approach significantly improves policy inference on tasks unsuccessful with the 
pre-trained policy, executing successfully: 57\% unseen background color tasks, 45\% unseen object tasks, 28\% of unseen color tasks, and 27\% when combining unseen objects and unseen backgrounds.

\section{Related Work} \label{relatedwork}
Integrating visual perception with robotic manipulation has been a goal for many decades. As deep learning techniques have matured over the last decade, they have increasingly been applied to visuo-motor learning in robotic systems \cite{finn2016,zhang2017}. Given the nature of these algorithms and the variance of visual perception, large-scale training data is required to yield generalised task execution \cite{levine2018,rt2}.

\subsubsection{Visual-Language Policies for Robotics}

Unification of the visual and language domains have yielded interesting results, driven by the progress in visual question answering~\cite{vqa}.
The fusion of visual-language models with visual-control networks enables language driven policies, as seen in CLIPort~\citeay{cliport}. The visual-language correlations learnt in CLIP~\citeay{clip} allows a language instruction to identify an object of interest in an image, which in conjunction with a policy network \citeay{transporternet}, performs tasks when an object of interest has been identified. Alternatively, combining several backbone models has been shown to successfully be trained to manipulate objects in vision-based simulation scenarios
\citeay{socratic}.

Our experiments deploy a pre-trained policy taken from CLIPort. Formally, it is a language and vision conditioned robotic policy $\pi$ which outputs an action $\mathbf{a}_t$ when given an input $\gamma_t = (\vist,\langt)$ consisting of a visual observation $\vist$ and a natural language instruction $\langt$. The action is a tuple $(\mathcal{T}_{pick}, \mathcal{T}_{place})$ where $\mathcal{T}_{pick}$ specifies the coordinates of the target object and $\mathcal{T}_{place}$ where it ought to be placed by the robotic manipulator. 

\subsubsection{Image Editing and Segmentation}
Recent advancements in text-to-image generation models, such as, DALL-E \citeay{dalle} and Stable Diffusion (SD)~\citeay{stablediffusion} demonstrate extraordinarily high-fidelity image generation via text descriptions. Works like InstructPix2Pix \citeay{instructpix2pix}  enable detailed semantic editing via the use of language instructions. Similar advancements have been seen in segmentation problems in computer vision. 
Following the success of DETR \citeay{detr} in end-to-end object detection, DINO \citeay{dino}, demonstrates emergent segmentation in the attention maps of Vision Transformers (ViT)~\cite{vit} under self-supervised learning. Following this work, Grounding DINO \citeay{groundingdino} utilized grounded pre-training 
to perform open-set object detection. 
In our work, we use Grounded-Segment Anything (GSAM) in combination with Stable Diffusion. GSAM combines Grounding DINO and Segment Anything (SAM) \citeay{segmentanything}, in which the SAM provides segmentation masks for all objects in the image.

\subsubsection{Semantically Imagined Experience}
Data augmentation is often used to overcome the cost of training visuomotor policies by altering previous experiences to produce novel training examples.
ROSIE \citeay{rosie} is a general and semantically-aware data augmentation strategy incorporating language prompts and generative models. 
It utilizes a user-provided language instruction in cooperation with a large-language model (LLM)\footnote{ROSIE's choice of LLM is GPT-3 \citeay{gpt3} but any LLM or even hand-engineered prompts can be used.} to produce three language prompts from the original language input: 
(i) a \emph{ViT region prompt} passed to a vision system (OWL-ViT \citeay{owlvit} in their case) in order to segment the region in which the robotic arm will be interacting e.g.~a drawer; (ii) a \emph{passthrough object prompt} passed also to OWL-ViT, but segmenting the robotic arm, which will be passing through the ViT region, to not alter this region with any text-to-image augmentation; and (iii) an \emph{inpainting prompt} passed to a text-to-image model for inpainting.
Imagen Editor~\citeay{imagen} (used in ROSIE) incorporates the inpainting prompt to alter the region identified by the ViT region prompt (ignoring the passthrough object prompt identified area) in a semantically meaningful manner, thereby creating a new visual observation that can be added to the training set.

\begin{figure*}[t]
    \centering
    \includegraphics[width=.98\linewidth]{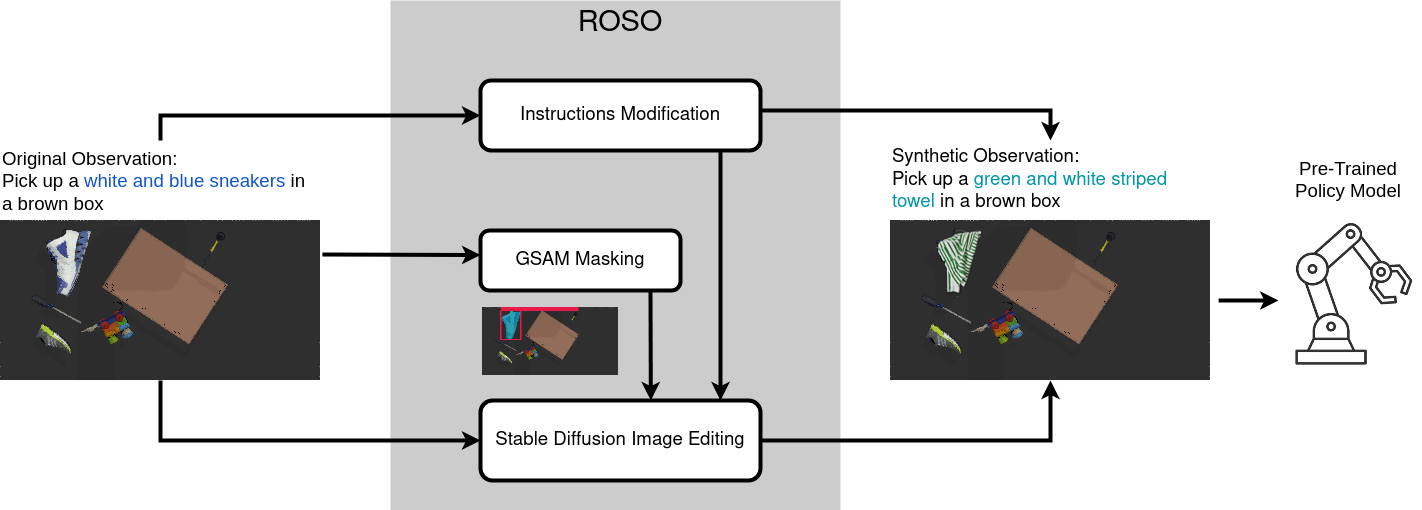}
    \vspace{-1mm}
    \caption{ROSO intercepts the observations $\gamma_t$ (prompt and image, on the left) to produce synthetic observations that are closer to previously seen ones, before feeding them into a pre-trained policy.}
    \label{fig:pipeline}
\end{figure*}

\section{Our Approach}
In this paper we propose that the performance of a pre-trained policy $\pi$ can be improved by synthetically altering its observations $\gamma_t$ \textbf{during inference time}, contrasting ROSIE's approach of simply diversifying training data.
We use a CLIPort trained policy $\pi$ to tackle tasks where in an environment $\varepsilon$ the agent obtains an observation $\gamma_t = (\vis, \lang)$ at time $t$, executes an action $\textbf{a}_t = \pi(\gamma_t)$, yielding the next observation and reward from the environment $(\gamma_{t+1},r_{t+1}) \gets \varepsilon(\textbf{a}_t)$. 

The policy $\policy$ (or in fact, any neural network) will perform best on inputs it has been trained on. Hence in ROSO we wish to alter an observation $\gamma_t$ which fails the task into an observation $\gamma_t'$ by altering $\gamma_t$ to look more like a training observation.

To demonstrate the improvement of $\pi$ via synthetic observations we first collect a dataset $\mathcal{D}_\textrm{unsucessful}$ of visual-language observations $\gamma_t$ in which the policy has failed its pick and place task, ie.~an observation for which most likely not enough data existed at training time. Unsuccessful tasks are defined by episode reward $r_{t+1} \neq$ 1. Hence, 
\[
\mathcal{D}_\textrm{unsuccessful} = \left\{ \gamma_t \mid \varepsilon(\pi(\gamma_t)) = (\gamma_{t+1},r_{t+1}) \textrm{ and } r_{t+1} \neq 1 \right\}.
\]

\subsection{Pipeline Overview}
\begin{algorithm}[b!]
\caption{Robotic Policy Inference via Synthetic Observations (ROSO)}\label{alg:cap}
\begin{algorithmic}
\Require $\gamma_t = (\vis, \lang) \in \mathcal{D}_\textrm{unsuccessful}$ 

\State $\mu \gets \textrm{GSAM}(\vis, \lang)$   \Comment{Acquire mask $\mu$}
\State $\lang' \gets \textrm{Prompt Mod}(\lang)$    \Comment{Semantically alter $\lang$}
\State $\vis' \gets \textrm{SD}(\mu, \lang', \vis)$     \Comment{Inpaint region $\mu(\vis)$}
\State $\textbf{a}_t' \gets \pi(\vis', \lang')$ \Comment{Apply CLIPort policy $\pi$}
\State $r_{t+1}' \gets \varepsilon(\textbf{a}_t')$ \Comment{Retrieve new reward $r_{t+1}'$}
\end{algorithmic}
\end{algorithm}

Our approach, \textbf{Ro}botic Policy Inference via \textbf{S}ynthetic \textbf{O}bservations (\textbf{ROSO}), uses Stable Diffusion to pre-process a robot's perception of novel objects during inference time to fit within its distribution of observations of the pre-trained policies.

We obtain the initial language instruction $\lang$ and visual observation $\vis$ to produce automated object masking (GSAM herein) before $\lang$ and $\vis$ are fed into $\policy$ (Figure~\ref{fig:pipeline} \&  Algorithm~\ref{alg:cap}). The target pick/place object in the visual observation $\vis$ is then altered to a new object $\vis'$ utilising masking and inpainting. 
Our aim is to replace a target that the CLIPort trained policy has never interacted with in training time, with one it has interacted with. For example, if the policy was trained to pick and place red cubes but had never been trained on blue cubes, upon encountering a blue cube the masking would identify the novel cube and inpainting would replace it with a red cube of the same dimensions.
In addition, we alter the initial language instruction $\lang$ into an altered instruction $\lang'$ which now makes reference to the inpainted object in the synthetic observation $\vis'$, so that the policy is able to execute successfully.
The synthetic observations creation by our pipeline (ROSO) can be split into two aspects: instruction modification and image modification.

\subsection{Instruction modification}\label{promptmod}
\begin{table*}[htp]
\centering
\begin{tabular}{ |>{\centering\arraybackslash}m{13em}|>{\centering\arraybackslash}m{16em}|>{\centering\arraybackslash}m{17em}| }
    \hline
    \textbf{Instruction Modification
    } & \textbf{Original} & \textbf{Modified}\\
    \hline
    \multirow{2}{*}{Color} & pick the \textcolor{orange}{orange} block in a \textcolor{red}{red} bowl & pick the \textcolor{brown}{brown} block in a \textcolor{red}{red} bowl\\
    \cline{2-3}
    & pick the \textcolor{Rhodamine}{pink} block in a \textcolor{purple}{purple} bowl & pick the \textcolor{ForestGreen}{green} block in a \textcolor{ForestGreen}{green} bowl\\
    \hline
    \multirow{2}{*}{Semantic Meaning} & pick the \textcolor{orange}{lion figure} in a brown box & pick the \textcolor{gray}{rhino figure} in a brown box\\
    \cline{2-3}
    & pick the \textcolor{RoyalPurple}{screwdriver} in a brown box & pick the \textcolor{NavyBlue}{scissors} in a brown box\\
    \hline
    \multirow{2}{*}{Image Edit Quality} & pick the \textcolor{orange}{lion figure} in a brown box & pick the \textcolor{ForestGreen}{green and white striped towel} in a brown box\\
    \cline{2-3}
    & pick the \textcolor{RoyalPurple}{screwdriver} in a brown box & pick the \textcolor{ForestGreen}{green and white striped towel} in a brown box\\
    \hline
\end{tabular}
\caption{Examples of modified prompts by the various instruction modification methods.}
\label{tbl:instruction_mods}
\end{table*}

\subsubsection{Color modification of the object}
Given the task of picking a colored block into a colored bowl, we observed that even though the concept of color has been exposed to CLIPort trained policies through CLIP, the policies fail to perform pick and place task on certain colors.  
To map unseen colors to seen colors, a colormap is obtained by running trained policies (see Appendix \ref{appendixc} for details).

\subsubsection{Modification Based on Semantic Meaning}
\begin{figure}[htp]
    \centering
    \includegraphics[width=\linewidth]{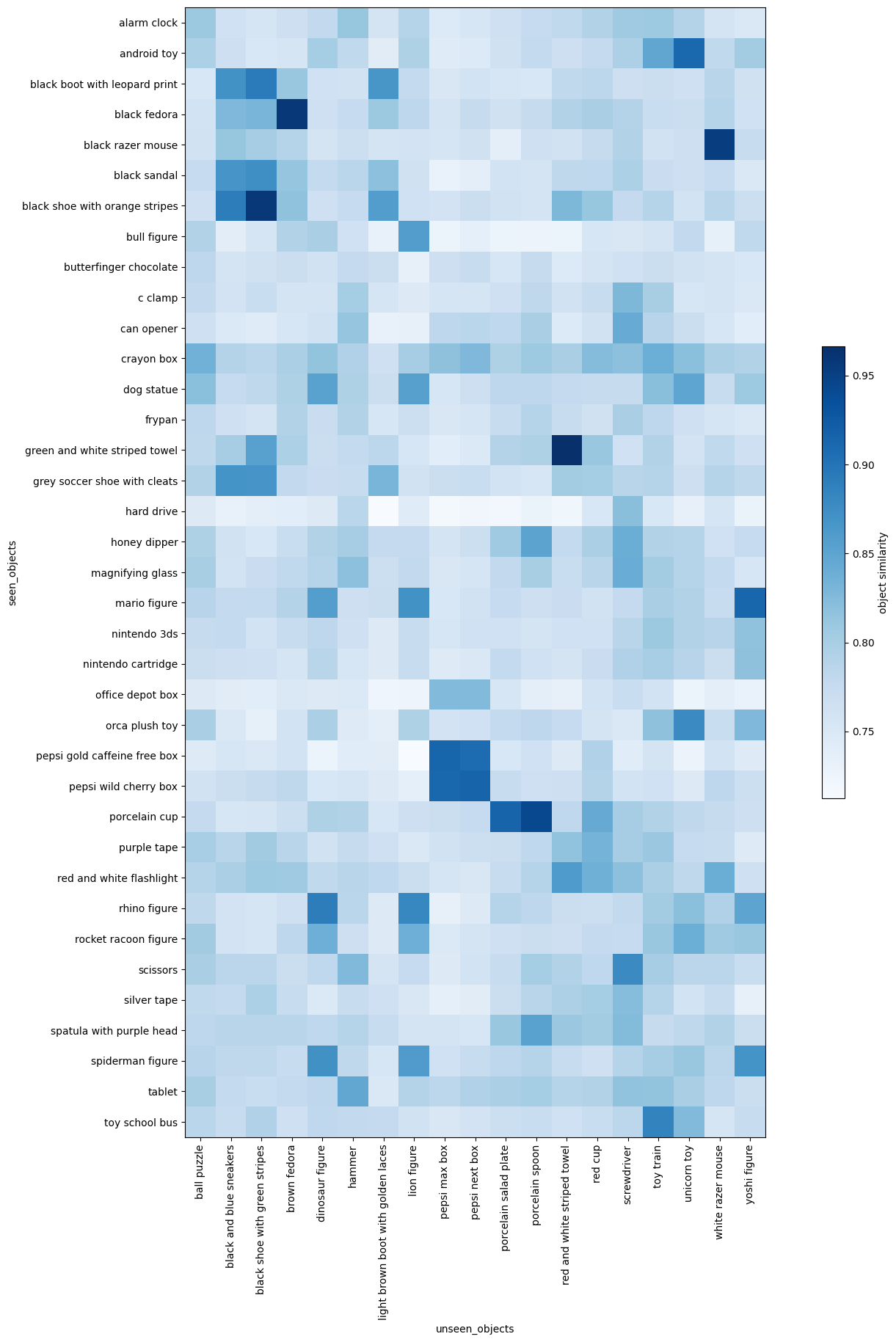}
    \vspace{-6mm}
    \caption{heatmap showing the semantic similarity between unseen objects and seen objects}
    \label{fig:OpenAI_ada_mapping}
\end{figure}

We can chose to replace the novel language instruction $\lang$ with a language instruction which was part of CLIPorts training set. From this set of pre-trained language instructions we choose the one which has the most semantic similarity to $\lang$. The semantic similarity between the language instructions $\lang$ and any other instruction $\ell$ from the CLIPort training set can be measured via the use of a text encoder\footnote{Our text encoder of choice is the Ada002 OpenAI embedding model. (\texttt{model text-embedding-ada-002} from \url{https://openai.com/blog/new-and-improved-embedding-model})} $\mathcal{E}$ and the Euclidian inner product $\langle \cdot, \cdot \rangle$. Hence
\[\lang' = \arg \max_{\ell} \langle \mathcal{E}(\lang), \mathcal{E}(\ell)\rangle\]
using this method. The list of seen and unseen objects are taken from the tasks defined in CLIPort \citeay{cliport}. Naturally, it is possible to generate a heatmap showing the inner product similarity between unseen objects and seen objects (Figure \ref{fig:OpenAI_ada_mapping}).

\subsubsection{Modification Based on the Image Edit Quality} We further investigated the prompt modification in terms of the image edit quality of Stable Diffusion and its alignment with the training dataset of CLIPort. 
We applied various image similarity test -- MSE, SIME and FID -- across objects in the training dataset and SD-generated images to find the most appropriate object to incorporate into our synthetic observation ($\vis'$, $\lang'$). 

\subsection{Image Modification}

\begin{figure}[b!]
    \centering
    \includegraphics[width=\linewidth]{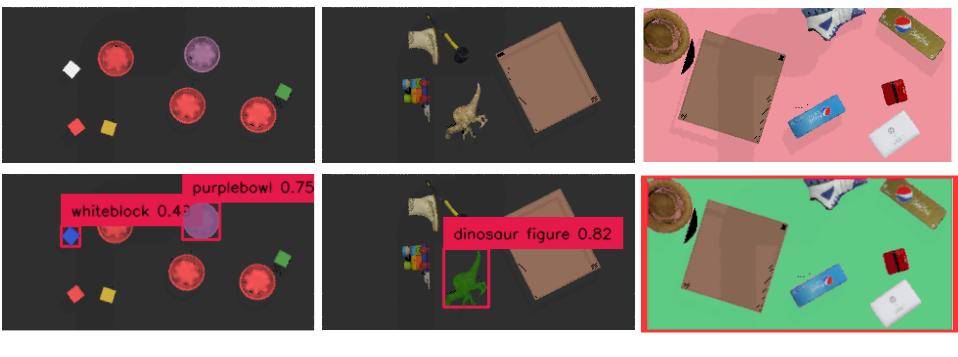}%
    \vspace{-2mm}
    \caption{Generalised segmentation methods (such as GSAM) can obtain a segmentation mask of unseen data (unseen color, novel objects, unseen background). The bottom row shows how these segmented parts can be changed by generative models.}
    \label{fig:GSAM_mask}
\end{figure}

\begin{figure}[t]
    \centering
    \includegraphics[width=\linewidth]{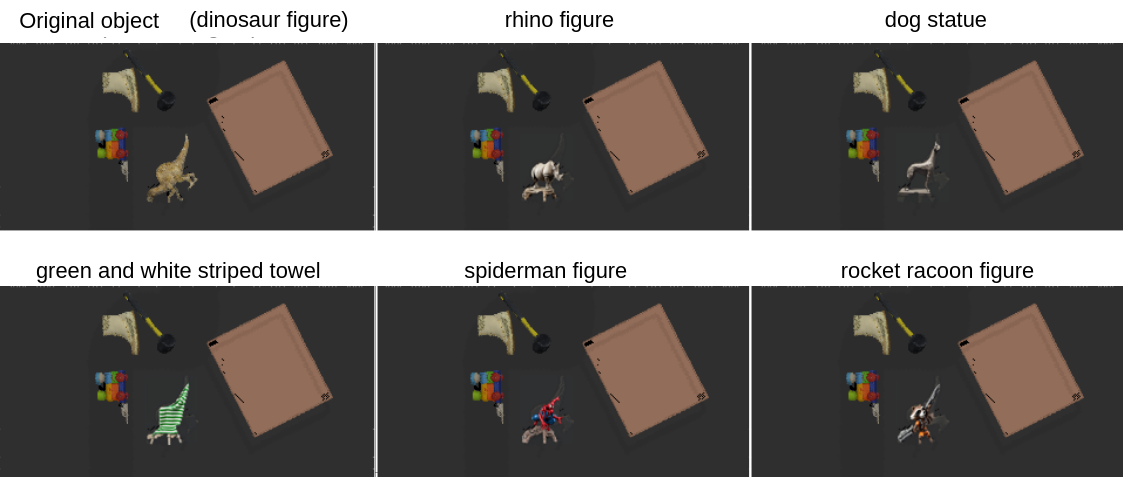}
    \vspace{-6mm}
    \caption{Examples of inpainting various seen objects (from the CLIPort training data) replacing an unseen objects (the dinosaur figure in the top-left image).}
    \label{fig:SD_objects}    
\end{figure}

To alter the visual observation $\vis$ and replace the unseen objects with the seen objects, we first need to identify the object of interest in $\vis$. We perform the identification step using GSAM to detect the object of interest and segment it to generate a mask $\mu$ (Figure~\ref{fig:GSAM_mask}). The masked region-of-interest can then be altered via a text-to-image diffusion model. Our approach involves changing part of the input image such as the object and background. Therefore, we specifically use Stable Diffusion (SD) inpainting, a latent text-to-image diffusion model capable of generating photo-realistic images given a text prompt, with the extra capability of inpainting the pictures by employing a mask (Figure~\ref{fig:SD_objects}). Our approach is agnostic to the in-painting model used. 

Moreover, ROSO only modifies the visual appearance of the object using inpainting. The original shape is preserved via the object mask and the depth is preserved using the depth map obtained from the original observation.

\section{Experiments \& Results}

\begin{figure*}[t!]
    \centering
    \includegraphics[width=\linewidth]{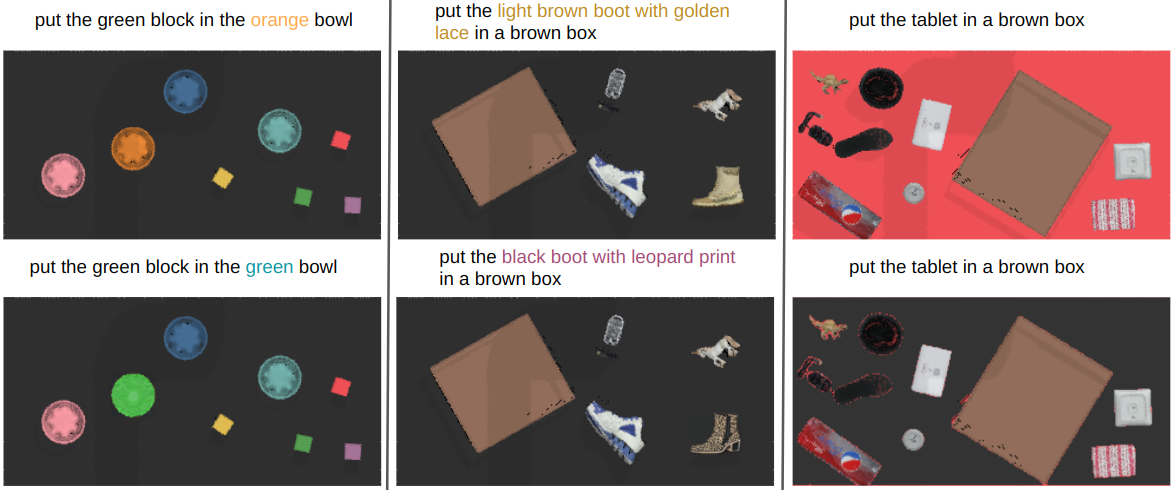}
    \vspace{-6mm}
    \caption{Example vision-language tasks (top) and synthetically altered version of the tasks by our pipeline (bottom)}
    \label{fig:tasks_and_their_edits}
\end{figure*}

The tasks we use for our experimental validation closely resemble the language-guided visual manipulation challenges found in previous work. Particularly, we compare the performance between a baseline (the pre-trained policy $\policy$ taken from CLIPort~\citeay{cliport}) and ROSO (Figure~\ref{fig:tasks_and_their_edits}).
These tasks come in two variants: ``seen'' and ``unseen'', depending on whether they include objects and scenarios that have not been provided in the training set.
For instance, when it comes to colors, the ``seen'' colors include {yellow, brown, gray, cyan}, while the ``unseen" colors include {orange, purple, pink, white}   to seen colors. There are also 3 colors, namely {red, green, blue}, that overlap and are used in both the ``seen'' and ``unseen'' variants.
In the case of packing objects, CLIPort utilizes a pool of 56 tabletop objects from the Google Scanned Objects dataset~\citeay{GoogleScannedObjects}
, which are divided into 37 objects seen during training and 19 objects that remain unseen until testing. For more details see the Appendix.

\subsubsection{Environment Setup}
We conduct all experiments on PyBullet using a Universal Robot UR5e equipped with a suction gripper. PyBullet simulations provide consistent and repeatable evaluation of our policy $\pi$ within diverse environments. The input observation $\vis$ is a top-down RGB-D reconstruction obtained from three strategically positioned cameras (each 640×480 pixels) encircling a rectangular table: one at the front, one on the left shoulder, and one on the right shoulder, all directed towards the center.

Our experimental evaluation captures empirical data on the improvement of a pre-trained policy's performance on unseen data.
We design our experiments, aiming to answer the following research questions:
\begin{enumerate}
    \item Can we improve a robotic policy's performance in the unseen domain by using generative AI to synthetically alter their observations into the seen domain during inference time?
    \item What are ways to improve synthetic observation creation, including prompt modification, automated masking and image editing?
    \item What are the different modes of failure that emerge when attempting to semantically alter a robotic policies experience?
\end{enumerate}

\begin{table*}[t]
\centering
{%
    \begin{tabular}{ 
    |p{14.4em}||p{8.75em}||p{8.75em}|p{13em}| }
    \hline
    \diagbox{Task}{Data Domain {\color{red}x}}&
        Seen Data \newline\small{\citeay{cliport}}&%
        Unseen Data \newline\small{\citeay{cliport}}&%
        \textbf{Unseen Data + Synthetic Observations} [Ours]
        \\
    \hline
    \hline
    put the {\color{red}x} block in a {\color{red}x} bowl & 30 failed runs & 57 failed runs & 41 failed (28\% more success) \\
    \hline
    pack an {\color{red}x} object in a brown box & 13 failed runs & 29 failed runs & 16 failed (45\% more success) \\
    \hline
    pack an object in a brown box,\newline{\color{red}x} background color & 13 failed runs & 91 failed runs & 39 failed (57\% more success)\\
    \hline
    pack an {\color{red}x} object in a brown box, {\color{red}x} background color & 13 failed runs & 96 failed runs & 70 failed (27\% more success)\\
    \hline
    \end{tabular}}
    \caption{100 variations are run for each task (where {\color{red}x} is varied). The number of failed runs significantly drops when using our synthetic data during inference, approaching the success rate on the training set (first column).}
    \label{tbl:hundred_demos}
\end{table*}

\subsection{Adapting to Unseen Tasks}
Considering that ROSO's performance improvement is bounded by CLIPort's performance on seen data our experiments collectively address the research question of adapting to unseen tasks 
during inference time. 
\subsubsection{Color Adaptation}
For the task \emph{Put a Block in a Bowl} 
a CLIPort-trained policy results in 57 failures, out of 100 runs with unseen object colors (compared to 30 failed runs for seen object colors, ie.~data that has been seen during training). ROSO is able to increase the successful outcomes, and reduce the failed runs for unseen object colors to 41 (from 57 without using synthetic observations, a 25\% increase in performance, see Table~\ref{tbl:hundred_demos}) while demonstrating its ability to synthesize observations that alter the color of the object from unseen to seen. Analyzing the failure cases of ROSO in this task, we see that modifying the object color into the same color as a distractor object present in the scene results in confusing the policy and offers opportunity to further increase the capability in future work.

\subsubsection{Object Adaptation}
In the context of unseen object domain, the task \emph{Pack an object} is employed (Appendix \ref{appendixa}). 
Object modification provides flexibility in image editing, enabling us to transform unseen objects into seen ones based on their semantic meanings or image editing quality. Both approaches yield performance improvements 
on unseen data. Additionally, we discovered that transforming all unseen objects into a single object is particularly effective, provided that this object possesses high-quality image editing and aligns well with CLIPort's training dataset. 
\subsubsection{Unseen Background Color}
For the task of \emph{Pack an object}, we also experimented with background color modification (Appendix \ref{appendixa}). Although we only altered the background color and test on more complex textures or patterns are left for future research, we observed that the policy was significantly distracted by the presence of an unseen background color. ROSO can detect the background region and transform it back into CLIPort's training domain, specifically a plain dark gray color. GSAM performed well in this task, resulting in a substantial performance improvement of a CLIPort-trained policy when using our synthetic observation pipeline.

\subsubsection{Unseen Objects in Unseen Background Color}
Finally, we run experiments requiring ROSO to change both object and background color. 
This is the most complex case as the presence of unseen background can adversely affect detecting unseen objects, increasing its difficulty to render into seen objects and thus alter the observation.
These experiments resulted in the lowest success rate (yet still a 27\% improvement over CLIPort) when compared to the other tasks, due to the completely novel observations. The failure is also caused from the background affecting 
the policy's ability to infer the affordance. Future improvements may include an investigation into the appropriate editing order (editing the background first then the target or vice versa) and determining which edited background yields the best policy performance.

\subsection{Comparison of Editing Methods}

\subsubsection{Inpainting Using the Whole Image}
Inpainting using the whole image involves using the entire scene of the original image which incorporates global information into the masked area. 
We find this approach sometimes fails at producing inpainting with satisfactory fidelity due to the addition of non-local information into the inpainted area. 
Additionally, using the entire image for inpainting can impose limitations on the size of the mask and the number of pixels that SD can effectively edit. When the mask region is too small, we observe SD struggling to recreate an object accurately within.

\begin{figure}[b!]
    \centering%
    \includegraphics[width=\linewidth]{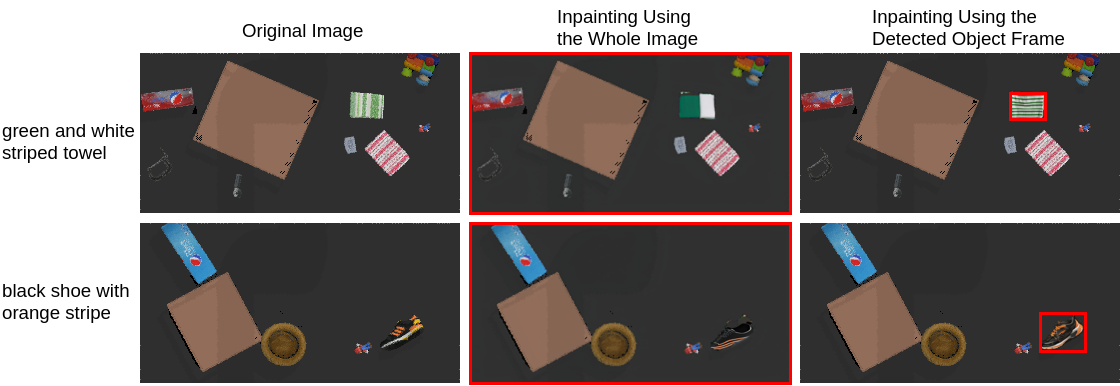}%
    \vspace{-1mm}
    \caption{Comparison between inpainting by passing the whole image into Stable Diffusion (middle column) vs passing the detected frame (right column). In each scenario, the regions inside the red squares were passed into SD to produce the observed results. Inpainting using the detected frame yielded higher-quality object generations.}
    \label{fig:inpaint_methods}
\end{figure}

\subsubsection{Inpainting Using the Detected Object Frame}
Once GSAM has detected an object and produced the object mask we consider the smallest frame (rectangle) which contains the detected mask, what we call the ``detected object frame''. 
Cropping this frame and passing it into SD for inpainting results in only the strictly local information around the detected object being utilized. We observe higher-quality edits using this method which requires minimal surrounding information, typically limited to background color or texture, making it a more accurate approach (Figure~\ref{fig:inpaint_methods}).

\subsubsection{Synthetic Object Editing Methods: Semantic Meaning vs Image Edit Quality}
To synthesize the object different methods are available\footnote{For ``put a block in a bowl'', editing methods A and B modify objects according to the color. For ``packing an object'', they modify objects based on the semantic meaning of their text and map to the most similar, seen object.}:
\begin{itemize}
    \item Method A: Inpaint mask with the whole image + modify based on color (put a block in a bowl) or semantic meaning (packing an object) of the object 
    \item Method B: Inpaint mask within its detected object frame + modify based on color or semantic meaning of the object 
    \item Method C: Inpaint mask within its detected object frame + modify based on the image editing quality
\end{itemize}

\begin{table}[t!]
\centering
{%
    \begin{tabular}{ |p{7em}||p{4em}|p{4em}|p{4em}| }
        \hline
        {\textbf{Task}} & \multicolumn{3}{c|}{\textbf{Method}} \\
         & A & B & C\\
        \hline
        Put a Block in a Bowl &  20\% & 22\% & \textbf{28\%} \\
        \hline
        Packing an Object & 16\% & 30\% & \textbf{45\%} \\
        \hline
    \end{tabular}}
    \caption{Success rate of ROSO for each editing method} 
    \label{tbl:ABC}
\end{table}

We observed that Method C, modifying objects based on the image editing quality, resulted in a higher success rate than the other methods (Table~\ref{tbl:ABC}). In particular, the object, ``Spider-man figure" achieved the highest success rate of 45\%.
While mapping unseen to seen object based on semantic meaning of the text may allow smoother transition from unseen to seen objects as their color or shape may be similar, it does not take into consideration the image quality of the object generated.
Furthermore, not all of the generated objects necessarily align with CLIPort's training dataset, despite both models incorporating the shared language component, CLIP, in their respective architectures.  
This misalignment is observed to impact task execution in some cases.

\subsection{Modes of Failure}

A baseline CLIPort-trained policy demonstrates a high success rate for seen data. However, its performance drops when dealing with unseen data, ie.~observations not experienced during training. This drop in success can be attributed to the policy's inability to accurately infer pick and place affordances or failures in grasping. With our approach, we aim to enhance the affordance inference of the policy for unseen objects and scenes.

\begin{figure}[b!]
    \centering
    \includegraphics[width=\linewidth]{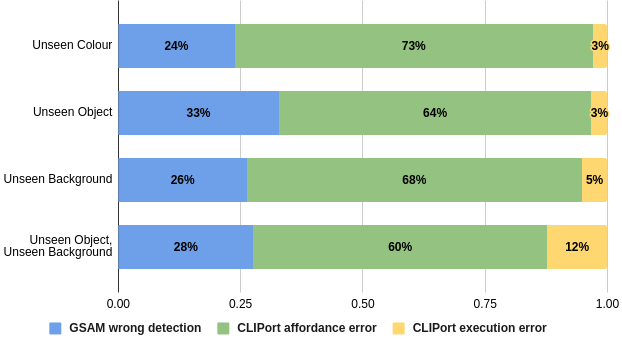}
    \vspace{-5mm}
    \caption{Ratio of failures in each component of ROSO. \emph{CLIPort affordance error} means CLIPort errors in pick and place prediction, which can be caused by SD image editing. \emph{CLIPort execution error} consists of the Robot failing to grab the object even though the pick and place affordance is correct}
    \label{fig:mode_of_failure}    
\end{figure}

While our approach does lead to performance improvements through the incorporation of synthetic observation (Table~\ref{tbl:hundred_demos}), it still exhibits failures (Figure~\ref{fig:mode_of_failure}). When analyzing these failures, we observed that they can be further categorized based on what stage of ROSO they correlate with. 

In the masking component, there are two primary types of failures: (i) {GSAM failing to detect prompted objects} and (ii) {GSAM incorrectly identifying objects}. The detection threshold parameter in GSAM can be changed to adjust the balance between these errors, whether increasing the chance to detect an object or to increase the accuracy to detect the correct object. However, the parameter merely changes the ratio of these two types of errors and they still contribute to the overall failure in the GSAM masking component.

On the other hand, the image editing exhibits most failures in our pipeline. The success of this component in ROSO is adversely affected when instruction modification requires the generation of complex objects through SD, resulting in {poor image quality} that impedes the policy's ability to find the correct affordance. Similarly, high-quality images produced by {SD may not align with the training dataset used by CLIPort}, further impacting ROSO's performance. As both cases result in the same execution error it is non-trivial to differentiate between the cases.

\subsubsection{Stable Diffusion and CLIPort Training Data Alignment}
To assess the alignment between SD and CLIPort's training data we test a range of metrics -- Mean Squared Error (MSE), Structural Similarity (SSIM), and Frechet Inception Distance (FID). MSE compares pixel-wise differences, SSIM measures pixel similarity, and FID considers semantic feature differences. 
However, we did not find a significant correlation between any of these metrics and the overall task success rate. Factors such as mask shape, variance in objects generated from SD, and SD's tendency to generate objects vertically within the mask may contribute to this lack of correlation.

\subsubsection{Consistency in Generated Objects}
Consistency in the generated object from SD also influenced the CLIPort success rate. For instance, inpainting an object as ``Pepsi wild cherry box" could result in various images such as {P}epsi cherry can, a Pepsi cherry box, or a cherry and Pepsi logo together, leading to unstable performance on CLIPort. Future work may investigate how to improve this by generating more complex prompts.

\subsubsection{Generating Objects Into Complex Mask Shapes}
The success rate can be attributed to the nature of the edited object. For instance, transforming an object into a ``green and white striped towel" is effective since a green and white striped towel is identifiable more from it's color and texture, making it more shape invariant. Shape invariant objects are easier for SD to generate. However, challenges arise when dealing with unseen objects featuring complex shapes, as SD may struggle to create an image that precisely matches the shape of the provided mask. Additionally, it is important to note that SD defaults to generating vertically oriented objects, which can lead to perplexing image edits when the unseen object is not oriented vertically.

\section{Conclusions}
We propose ROSO, a novel approach to generate synthetic observations for enhancing pre-trained visuomotor policies and successfully execute unseen tasks. 
We show in our experiments that using generative AI to synthetically augment a robotic policies observations is a viable alternative to additional training.
When tested against tasks involving unseen scenarios from the CLIPort dataset,
our ROSO pipeline improves the success rate by up to 57\%. It may provide a more efficient approach for improving robotic task execution in zero-shot scenarios. 

To further enhance robotics with ROSO 
it is imperative to consider various facets. For robotic manipulation, these include the physical object properties and gripper limitations, distinguishing between seen and unseen tasks, managing sequential tasks, and synthesizing observations seamlessly across continuous frames.

\bibliography{bibliography}
\bibliographystyle{named}

\section*{Appendix and Additional Material}
\renewcommand{\thesubsection}{\Alph{subsection}}
\subsection{Task details}
\label{appendixa}
The tasks are defined in the CLIPort paper~\citeay{cliport}, Appendix A and we extended them to challenge our trained policy on unseen color, unseen object and unseen background color.

\begin{figure}[b!]
    \centering
    \includegraphics[width=\linewidth]{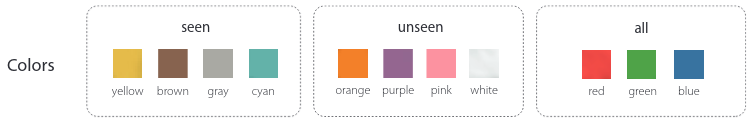}
    \includegraphics[width=\linewidth]{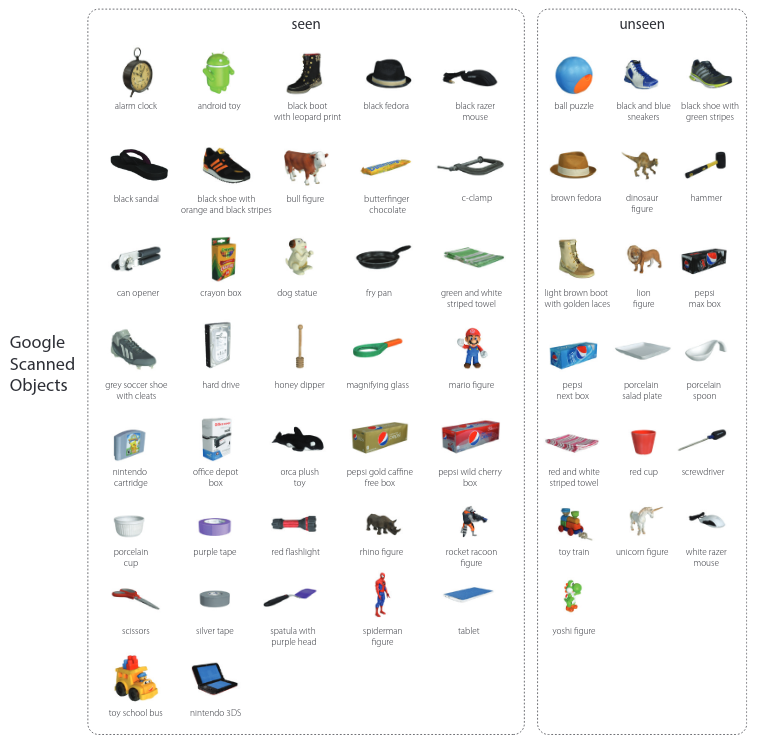}
    \vspace{-4mm}
    \caption{Attributes and objects across seen and unseen splits defined in CLIPort}
    \label{fig:task_description}
\end{figure}

\subsubsection{Task: Put a Block in a Bowl}
\textbf{Description:} Place the block of a specified color in a bowl of specified color. Each task instance contains several distractor blocks and bowls with randomized colors except for the ones used to plick and place. 
This task does not require precise placements and mostly tests an agent’s ability to ground color attributes\\
\textbf{Objects:} \emph{Put a Block in a Bowl Seen Colors} is trained and evaluated on [seen, all] colors combined from Figure \ref{fig:task_description} for both blocks and bowls. \emph{Put a Block in a Bowl Unseen Colors} is trained on seen colors but evaluated on [seen, unseen, all] colors combined from Figure \ref{fig:task_description} for both block and bowl\\
\textbf{Success Metric:} The block of the specified color is within the bounds a bowl of the specified color. The final score is defined by whether the correct block is in the correct bowl in the scene

\subsubsection{Task: Packing an Object}

\textbf{Description:} Place the specified object in the brown box. This task does not require precise placements or following a specific action sequence. The task cannot be solved by counting the number of objects since there are distractor objects and each scene contains objects of multiple categories.\\
\textbf{Objects:} \emph{Pack a seen object} is trained and evaluated on all 56 objects, whereas 
\emph{Pack an unseen object} is trained on 37 seen objects but evaluated on 19 unseen objects in Figure~\ref{fig:task_description}.\\
\textbf{Success Metric:} The specified object of a category are within the bounds of the brown box. The final score is the total volume of the correct objects in the box, divided by the total volume of the relevant objects of the specified category in the scene.

\subsubsection{Task: Packing an Object in Unseen Background Color}
The task resemble the task ``Packing an Object".  Task, Objects and Success Metric are the same except the initial ``darkgrey" background color is changed randomly to test CLIPort's robustness against different background colors.

\subsection{Implementation details}
We ran ROSO and observation alteration during inference time and with less compute compared to ROSIE.

CLIPort baseline and CLIPort + ROSO are run on local environment with consumer GPU, RTX2070 and generation of synthetic images is done on google colab free tier, T4 GPU as it required around 14G of RAM memory space.

Firstly, failed CLIPort tasks are collect, then they are processed on google colab for masking, prompt modification and image editing. Finally, the trained polciy runs again with the synthetic observations in the local environment. NB: the pipeline is run automatically without human involvement, other than transferring files between local environment and google colab which is done manually for this paper.

\subsection{Instruction Modification}
\label{appendixc}
In this section, we further explain the mapping between unseen data to seen data in detail.

\subsubsection{Modiciation based on the color of the object}
\begin{figure}[b!]
    \centering
    \includegraphics[width=0.9\linewidth]{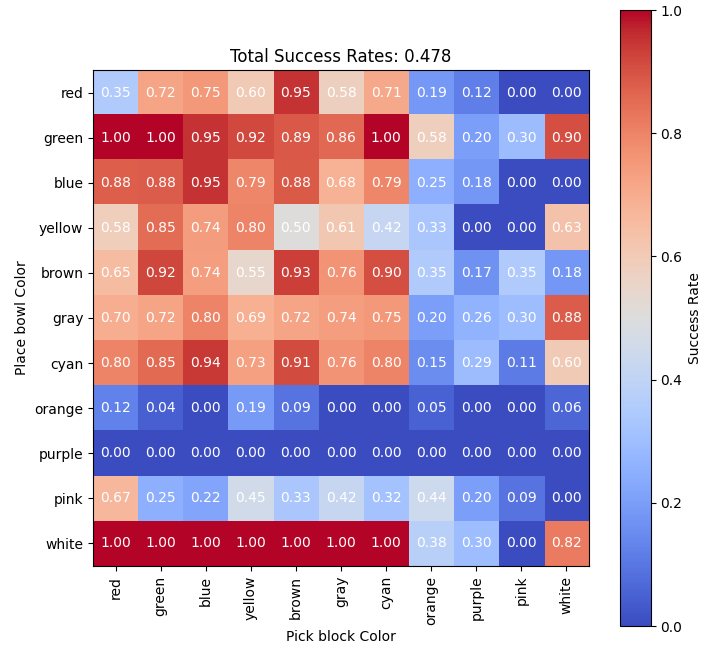}%
    \vspace{-2mm}
    \caption{colormap illustrating the CLIPort success rate for each pick and place combination}
    \label{fig:colormap}
\end{figure}

During the run to collect failed demos in CLIPort, a colormap (Figure \ref{fig:colormap}) is generated to illustrate the success rates for different combinations of pick and place object colors. It's important to note that this color map is not symmetrical and can be utilised during the inference phase of CLIPort to enhance its performance. For instance, consider the prompt ``pick the orange block in a red bowl," which involves the unseen color orange. In this case, selecting the brown color for picking yields the highest success rate given place color is red. Consequently, the color orange in the prompt is modified to brown.

\vspace{30mm}
\subsubsection{Modification Based on The Semantic Meaning of the Object}

\end{document}